\documentclass[sigconf]{acmart}
\AtBeginDocument{%
  }

\copyrightyear{2026}
\acmYear{2026}
\setcopyright{cc}
\setcctype{by}
\acmConference[KDD '26]{Proceedings of the 32nd ACM SIGKDD Conference on Knowledge Discovery and Data Mining V.1}{August 09--13, 2026}{Jeju Island, Republic of Korea}
\acmBooktitle{Proceedings of the 32nd ACM SIGKDD Conference on Knowledge Discovery and Data Mining V.1 (KDD '26), August 09--13, 2026, Jeju Island, Republic of Korea}
\acmPrice{}
\acmDOI{10.1145/3770854.3780211}
\acmISBN{979-8-4007-2258-5/2026/08}

\setcopyright{none}
\renewcommand\footnotetextcopyrightpermission[1]{}
\usepackage{newfloat}
\usepackage{listings}
\usepackage{algorithm}
\usepackage{algorithmic}
\usepackage{adjustbox}
\usepackage{color}
\usepackage{xcolor}
\usepackage{colortbl}
\usepackage{array}
\usepackage{multirow}
\usepackage{booktabs}
\usepackage{amsmath,amsfonts}

\usepackage{bm}
\usepackage{enumitem}
\usepackage[most]{tcolorbox}
\usepackage{float}
\usepackage{makecell}
\usepackage{subcaption}
\usepackage{xspace}

\usepackage{hyperref}
\hypersetup{
    hypertex=true,
    colorlinks=true,
    linkcolor=blue,
    anchorcolor=blue,
    citecolor=blue
}

\usepackage{cleveref}

\newcommand{\NumBaseline}{10\xspace}
\newcommand{\M}{RADAR\xspace}
\newcommand{\task}{FNVD\xspace}

\newcommand{\bc}[1]{\textcolor{blue}{#1}}

\NewDocumentCommand{\rp}{O{comment} m}{%
  \IfEqCase{#1}{%
    {comment}{\textcolor{red}{\textbf{[@RP COMMENT: #2]}}}%
    {done}{\textcolor{green}{\textbf{@RP DONE: [#2]}}}%
    {todo}{\textcolor{blue}{\textbf{@RP TODO: [#2]}}}%
  }[\textcolor{gray}{#2}]%
}

\usepackage{amsthm}

\newcommand{\vx}{\vec{x}}
\newcommand{\vy}{\vec{y}}

\begin{document}

\title[Nip Rumors in the Bud: Retrieval-Guided Topic-Level Adaptation for Test-Time Fake News Video Detection]{
Nip Rumors in the Bud: Retrieval-Guided Topic-Level Adaptation for Test-Time Fake News Video Detection
}


\author{Jian Lang}
\email{jian_lang@std.uestc.edu.cn}
\orcid{0009-0009-0876-0497}
\affiliation{%
  \institution{University of Electronic Science and Technology of China}
  \city{Chengdu}
  \state{Sichuan}
  \country{China}
}

\author{Rongpei Hong}
\email{rongpei.hong@std.uestc.edu.cn}
\orcid{0009-0007-4977-1657}
\affiliation{%
  \institution{University of Electronic Science and Technology of China}
  \city{Chengdu}
  \state{Sichuan}
  \country{China}
}

\author{Ting Zhong}
\email{zhongting@uestc.edu.cn}
\orcid{0000-0002-8163-3146}
\affiliation{%
  \institution{University of Electronic Science and Technology of China}
  \city{Chengdu}
  \state{Sichuan}
  \country{China}}

\author{Yong Wang}
\orcid{0000-0002-8699-8355}
\email{wangyong@ipplus360.com}
\affiliation{%
  \institution{Aiwen Technology Co., Ltd.}
  \city{Zhengzhou}
  \state{Henan}
  \country{China}}

\author{Fan Zhou}
\authornote{Corresponding author.}
\email{fan.zhou@uestc.edu.cn}
\orcid{0000-0002-8038-8150}
\affiliation{%
  \institution{University of Electronic Science and Technology of China}
  \city{Chengdu}
  \state{Sichuan}
  \country{China}
}
\affiliation{%
  \institution{Intelligent Digital Media Technology Key Laboratory of Sichuan Province}
  \city{Chengdu}
  \state{Sichuan}
  \country{China}
}

\renewcommand{\shortauthors}{Jian Lang et al.}

\begin{abstract}
Fake News Video Detection (\task) is critical for social stability.
Existing methods typically assume consistent news topic distribution between training and test phases, failing to detect fake news videos tied to emerging events and unseen topics.
To bridge this gap, we introduce \textbf{\M}, the first framework that enables test-time adaptation to unseen news videos.
\M pioneers a new \textit{retrieval-guided adaptation paradigm} that leverages stable (source-close) videos from the target domain to guide robust adaptation of semantically related but unstable instances.
Specifically, we  propose an \textit{Entropy Selection-Based Retrieval mechanism} that provides videos with stable (low-entropy), relevant references for adaptation.
We also introduce a \textit{Stable Anchor-Guided Alignment module} that \textit{explicitly} aligns unstable instances' representations to the source domain via distribution-level matching with their stable references, mitigating severe domain discrepancies.
Finally, our novel \textit{Target-Domain Aware Self-Training paradigm} can generate informative pseudo-labels augmented by stable references, capturing varying and imbalanced category distributions in the target domain and enabling \M to adapt to the fast-changing label distributions.
Extensive experiments demonstrate that \M achieves superior performance for test-time \task, enabling strong on-the-fly adaptation to unseen fake news video topics.
\end{abstract}




\maketitle

\section{Introduction}
\label{sec:intro}

\begin{figure}[t]
    \begin{subfigure}[b]{0.99\columnwidth}
  \centering
  \includegraphics[width=0.8\columnwidth]{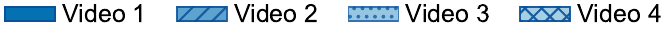}
  \end{subfigure}
  \centering
  \begin{subfigure}[b]{0.495\columnwidth}
  \centering
  \includegraphics[width=\columnwidth]{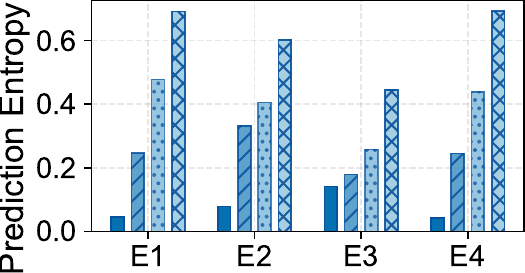}
  \vspace{-5mm}
  \caption{From FakeTT to FVC.}
  \end{subfigure}
  \hfill
  \begin{subfigure}[b]{0.495\columnwidth}
  \centering
  \includegraphics[width=\columnwidth]{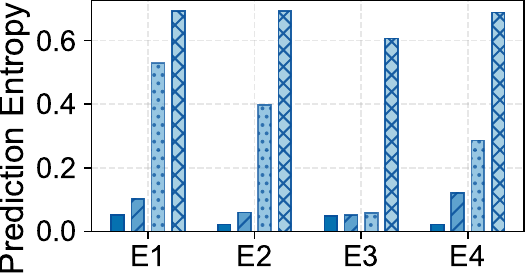}
  \vspace{-5mm}
  \caption{From FVC to FakeTT.}
  \end{subfigure}
    \vspace{-7mm}
  \caption{Prediction entropy of source model on four sampled news videos from four events (E1 --- E4) on the target dataset.}
  \vspace{-4mm}

  \label{fig:exp-intro}
\end{figure}

Popular video-sharing platforms like YouTube and TikTok have become key sources for the latest news. 
\textit{However}, their open nature also attracts malicious actors who spread deceptive content through videos, often laced with misleading rumors or false information. This poses serious risks to viewers' safety and erodes public trust~\cite{bu2023combating}. 
As a result, developing effective Fake News Video Detection (\task) methods is crucial and carries significant real-world implications.

Current \task approaches typically 
model multimodal information --- such as vision, text, and audio --- to uncover deceptive clues~\cite{choi2021using, qi2023fakesv, hong2025following}.
For instance, SV-FEND~\cite{qi2023fakesv} captured cross-modal correlations to select informative features and leveraged comments for detection.
Yet, fake news videos evolve quickly, often linked to breaking events, creating substantial topic-level distribution shifts between historical training videos and new instances~\cite{zafarani2019fake, choi2021using}.
For example, fake news videos about COVID-19 pandemic often contain medical misinformation and false health advice~\cite{nakov2021fake}, while those concerning the Russia-Ukraine conflict disseminate fabricated battlefield reports and geopolitical narratives~\cite{la2023retrieving} --- exhibiting substantial differences in both form and content.
Unfortunately, existing methods assuming \textit{consistent} topic distributions during training and test, when trained on outdated data, fail to handle those out-of-distribution samples from emerging events.

One simple solution is to annotate these novel new videos for finetuning~\cite{chen2025adaptation}. 
But this undermines the goal of detecting fake news at its earliest stage, rather than relying on hindsight after harm has occurred.
Unsupervised domain adaptation (UDA)~\cite{ganin2017domainadversarial, hoffman2018cycada, xu2023unsupervised} seems promising, as it leverages training data (source domain) and emerging unlabeled instances (target domain) to learn domain-invariant representations for generalization.
\textit{However}, 
UDA requires access to source data for adaptation, 
which is often impractical in \task due to the heavy storage overhead and privacy concerns~\cite{zhang2018crowdsourcing}.
To overcome the issues, we advocate a more rigorous, practical setting, namely test-time \task: \textit{Pre-trained \task models must promptly adapt to the evolving news videos with novel topics and events, while without access either source-domain videos or target-domain labels}, aligning with fully test-time adaptation (TTA) principles.

Existing TTA methods primarily rely on entropy minimization (EM) and its variants as self-supervised strategies~\cite{grandvalet2004semi, press2024entropy}, linking high-confidence (low entropy) predictions to higher prediction accuracy, and iteratively reducing entropy on unseen test data for adaptation~\cite{wang2021tent, niu2022efficient, zhang2025come, guo2025smoothing}.
Despite their success, applying them to test-time FNVD is intractable for the following two reasons:

\noindent \textbf{Challenge (1): Substantial discrepancies in feature representations between source and target domains}.
Most TTA approaches assume mild distribution shift from synthetic corruptions (e.g., blur or noise) on source data~\cite{wang2021tent, niu2022efficient, zhao2023pitfalls}, which EM can handle by \textit{implicitly} aligning representations~\cite{liang2020really, press2024entropy}. 
In FNVD, \textit{however}, topic-level shifts cause severe representation gaps (cf. \Cref{subsec:prelim} for experimental evidence), rendering indirect alignment insufficient.

\noindent \textbf{Challenge (2): Imbalanced and fluctuating category distributions}. 
Public attention varies across events, leading fake news creators to flood hot topics videos with forgeries while ignoring others~\cite{broda2024misinformation}, and resulting in highly skewed and varying proportions of fake and real news videos in diverse events.
In \task datasets like FakeSV~\cite{qi2023fakesv}, FakeTT~\cite{bu2024fakingrecipe}, and FVC~\cite{papadopoulou2019corpus}, 85.5\%, 79.0\%, and 98.4\% of news events showcase extreme imbalances, with the category distribution of news videos associated with such events being highly skewed (majority:minority > 4:1).
\textit{Unfortunately}, as mentioned in~\cite{zhao2023pitfalls}, most TTA methods work best only when the class prior distribution (i.e., the relative proportions of different classes) in the target domain is \textit{stable} and \textit{matches} that of the source domain.
\textit{Consequently}, the imbalanced and ever-changing label distributions within each newly arriving news video cluster across diverse events pose a severe challenge for existing TTA methods.

\begin{figure}[t]
    \centering
    \includegraphics[width=0.99\columnwidth]{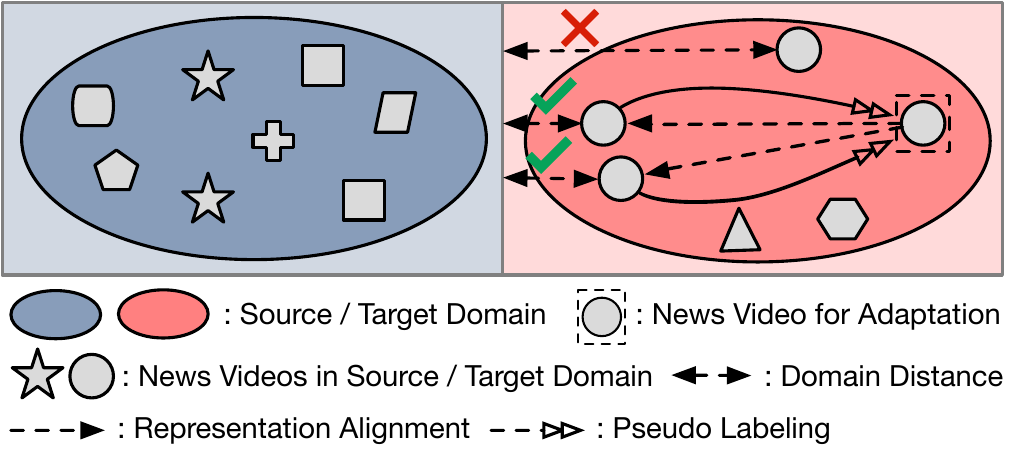}
    \vspace{-1mm}
    \caption{Concept diagram of our \M: the source-close news videos in the target domain are utilized to guide the robust test-time adaptation of the uncertain video instances. Instances with the same shape denote semantically relevant.}
    \label{fig:intro}
    \vspace{-6mm}
\end{figure}

To address these challenges, we observed that a source-trained model, without adaptation, produces low prediction entropy for some target-domain videos within the same event --- but high entropy for others. 
As presented in ~\Cref{fig:exp-intro}, when a source model is trained on one dataset and tested on the other (i.e., across FVC and FakeTT), it produces different levels of prediction entropy on four news videos from each sampled event.
The low-entropy videos often share common patterns with source samples (e.g., similar editing techniques or narratives), leading the source model to produce \textit{source-closer representations}~\cite{liang2020really, press2024entropy, gao2023back} and \textit{more accurate detections}~\cite{wang2021tent}. 
In light of these, a key question arises: \textit{Can we use these stable, low-entropy news videos to guide the test-time adaptation of unstable, high-entropy ones in the same or similar event?}

We answer affirmatively with \textbf{\M}, 
a novel \underline{\textbf{R}}etrieval-\underline{\textbf{A}}ugmented \underline{\textbf{D}}istribution \underline{\textbf{A}}lignment and target-aware self-t\underline{\textbf{R}}aining framework that, for the first time, enables TTA for \task.
As illustrated in \Cref{fig:intro}, \M pioneers a \textit{retrieval-guided adaptation paradigm} that leverages semantically relevant, stable target-domain videos to assist the effective adaptation of unstable instances.
Specifically, we first maintain a continually updated memory and introduce a new \textit{Entropy Selection-Based Retrieval} mechanism to fetch low-entropy, similar references (likely from the same event) from the memory for each query video. 
Since the source data is inaccessible, the retrieved stable target instances serve as \textit{source-domain proxies}, guiding robust adaptation.
To tackle \textbf{Challenge (1)}, our \textit{Stable Anchor-Guided Alignment} module \textit{explicitly} matches unstable instances' representations to their references at the distribution level, effectively bridging the source-target gaps. 
For \textbf{Challenge (2)}, a direct approach is to retrain the model on the newly arriving news video clusters that exhibit imbalanced category distributions. 
Inspired by this, our \textit{Target-Domain Aware Self-Training} paradigm moves a further step, assigning target news videos with pseudo-labels augmented by their low-entropy references.
These labels faithfully capture event-specific category distributions in the incoming video clusters, and are then utilized to self-train the model for rapid adaptation to imbalances.
Our main contributions include:
\begin{itemize} [leftmargin=*, topsep=2pt, partopsep=0pt, itemsep=0pt]

\item We are the first to recast the traditional \task task into a new research focus, namely \textbf{test-time \task}, where the pre-trained models are required to promptly adapt to novel fake news videos associated with emerging events and topics in an online manner, without accessing past training videos or any labels from incoming video streams (TTA setting).
We then elucidate the unique yet significant challenges in achieving 
test-time \task, and propose \textbf{\M}, which pioneers a novel retrieval-guided adaptation paradigm to enable effective TTA for \task.

\item A new \textit{Entropy Selection-Based Retrieval mechanism}, which provides each news video with stable (low-entropy) yet highly relevant reference instances to support robust adaptation.

\item A novel \textit{Stable Anchor-Guided Alignment module}, which explicitly bridges the substantial representation gaps caused by the topic-level distribution shift of the news videos between the source and target domains with the guidance of the stable references.

\item An effective \textit{Target-Domain Aware Self-Training paradigm}, which enables rapid adaptation to ever-changing target category distributions by self-training the model with pseudo-labels that accurately reflect the current label distribution in the target.

\end{itemize}

Extensive experiments on three video datasets under two proposed evaluation protocols for test-time \task showcase that \M achieves the remarkable performance compared to \NumBaseline baselines, offering a strong path for TTA in \task. 
The code and data are provided at 
\bc{\url{https://github.com/Jian-Lang/RADAR}}.

\section{Related Work}
\label{sec:related}

\subsection{Fake News Video Detection}
Fake News Video Detection (\task) aims to detect any misleading news content in online videos.
Early works in \task merely utilized single-modal information to assess the video authenticity~\cite{papadopoulou2019corpus, serrano2020nlpbased}, which may omit the potential deceptive clues in the unused modalities and lead to compromised detection.
Recent studies adopted various carefully designed multimodal approaches for enhanced \task~\cite{qi2023fakesv, bu2024fakingrecipe, hong2025following, li2025real, choi2021using, qi2023heads}.
For instance, SV-FEND~\cite{qi2023fakesv} leveraged cross-modal correlations and video comments for prediction.
REAL~\cite{li2025real} aligned the news videos
with category-specific prototypes to amplify subtle manipulation cues for enhanced detection.
\textit{However}, existing methods in \task often implicitly assume a consistent news topic distribution between the training and test phases, and thus struggle to adapt to the fake news videos associated with budding and unseen events.
\textit{Although} FANVM~\cite{choi2021using} attempted to learn topic-agnostic representations, its training is confined to limited event categories, which restricts its ability to generalize to unforeseen events.
\textit{Moreover}, it fails to consider the ever-changing and imbalanced category distributions in emerging news video streams.
In contrast, our proposed \textbf{\M} can adapt well to unseen news video instances on the fly at test time without being restricted to fixed training event categories.
\textit{Furthermore}, \M addresses the unique challenges (\textbf{substantial representation discrepancy} \& \textbf{unbalanced category distributions}) for test-time \task via a novel retrieval-guided adaptation paradigm, which facilitates the robust adaptation to the uncertain target-domain news videos by leveraging their stable and low-entropy references.

\subsection{Test-Time Adaptation}
Fully Test-Time Adaptation (TTA) focuses on adapting a model pre-trained on a source domain to a distribution-shifted target domain at test time, while assuming no access to labels of target data and the original source data~\cite{liang2020really, wang2021tent}.
Most studies in TTA build upon the principle of self-supervised Entropy Minimization (EM) and its variations (e.g., self-training~\cite{liang2020really}, efficient EM~\cite{niu2022efficient, niu2023towards}, marginal EM~\cite{zhang2022memo, shu2022test}). 
These approaches aim to minimize the prediction entropy of the model on the target data, thereby encouraging the model to produce confident and more accurate outputs on previously unseen samples.
For example, 
SAR~\cite{niu2023towards} jointly minimized the prediction entropy and the sharpness of entropy with batch-agnostic normalization optimization to tackle the wild TTA scenarios.
In contrast to single-modality scenarios, recent efforts explored applying TTA to more complex yet practical multimodal settings~\cite{yang2024testtime, guo2025smoothing, zhao2025attention}. 
For instance, SuMi~\cite{guo2025smoothing} smoothed the adaptation process and performed EM on samples with rich multimodal information.
\textit{However}, mainstream TTA works assumed a relatively mild distribution shift (i.e., corruptions on clean data) and thus relied on EM-based strategies to \textit{implicitly} align target representations with the source domain.
This indirect approach falls short in tackling the drastic representation gaps introduced by the topic-level distribution shift in \task.
\textit{Moreover}, existing TTA methods struggle to handle the imbalanced and ever-changing category distributions associated with different breaking news in \task~\cite{zhao2023pitfalls}, even SAR only had an initial attempt without an effective solution.

To remedy the lack of effective solutions for test-time \task, 
our framework \textbf{\M} \textit{explicitly} bridges the domain representation discrepancy by aligning the uncertain news videos to their low-entropy reference video instances.
It then endows these news videos with pseudo-labels augmented by their reference instances.
These pseudo-labels well reflect the label distributions of news video clusters associated with the respective breaking events and guide the model in adapting the new category distributions on the fly.

\section{Methodology}
\label{sec:methodology}

\newtheoremstyle{customremark}
  {0.3em}                
  {0.3em}                
  {\itshape}                
  {}                
  {\bfseries}       
  {.}               
  {0.3em}           
  {\thmname{#1}~\thmnumber{#2}} 

\theoremstyle{customremark}
\newtheorem{remark}{Remark}

\subsection{Preliminary}
\noindent \textbf{Definition of Test-Time \task.}
For a given news video $\mathcal{S}$ with its vision modality $v$ (i.e., video frames), text modality $t$ (i.e., video title and on-screen text), and audio modality $a$ (i.e., audio transcript), \task aims to determine whether $\mathcal{S}$ is \textit{real} or \textit{fake} by considering all the modalities~\cite{qi2023fakesv, li2025real}.
The detailed video feature extraction process is available in \Cref{app:feature}.
To coincide with TTA setting, we further define $\mathcal{F}_{\theta} = \{f_{{\theta}_{v}}, f_{{\theta}_{t}}, f_{{\theta}_{a}}, \mathcal{P}_{{\theta}}\}$ as the source model pre-trained on a labeled source-domain dataset $\mathcal{D}_{s} =\{\mathcal{S}_i\}_{i = 1}^{N_s} = \{(v_i, t_i, a_i, y_i)\}_{i = 1}^{N_s}$, where $f_{{\theta}_{v}}$, $f_{{\theta}_{t}}$, and $f_{{\theta}_{a}}$ denote the vision, text, and audio encoders, and $\mathcal{P}_{{\theta}}$ is the modal fusion and prediction network.
Test-time \task aims to improve the detection performance of the source model $\mathcal{F}_{\theta}$ on a distribution-shifted (incurred by the budding news topics) target-domain dataset $\mathcal{D}_{t} = \{\mathcal{S}_i\}_{i = 1}^{N_t} = \{(v_i, t_i, a_i)\}_{i = 1}^{N_t}$ on the fly at inference time, without accessing the labels and the source data.

\begin{figure*}[t]
  \centering
  \includegraphics[width=\linewidth]{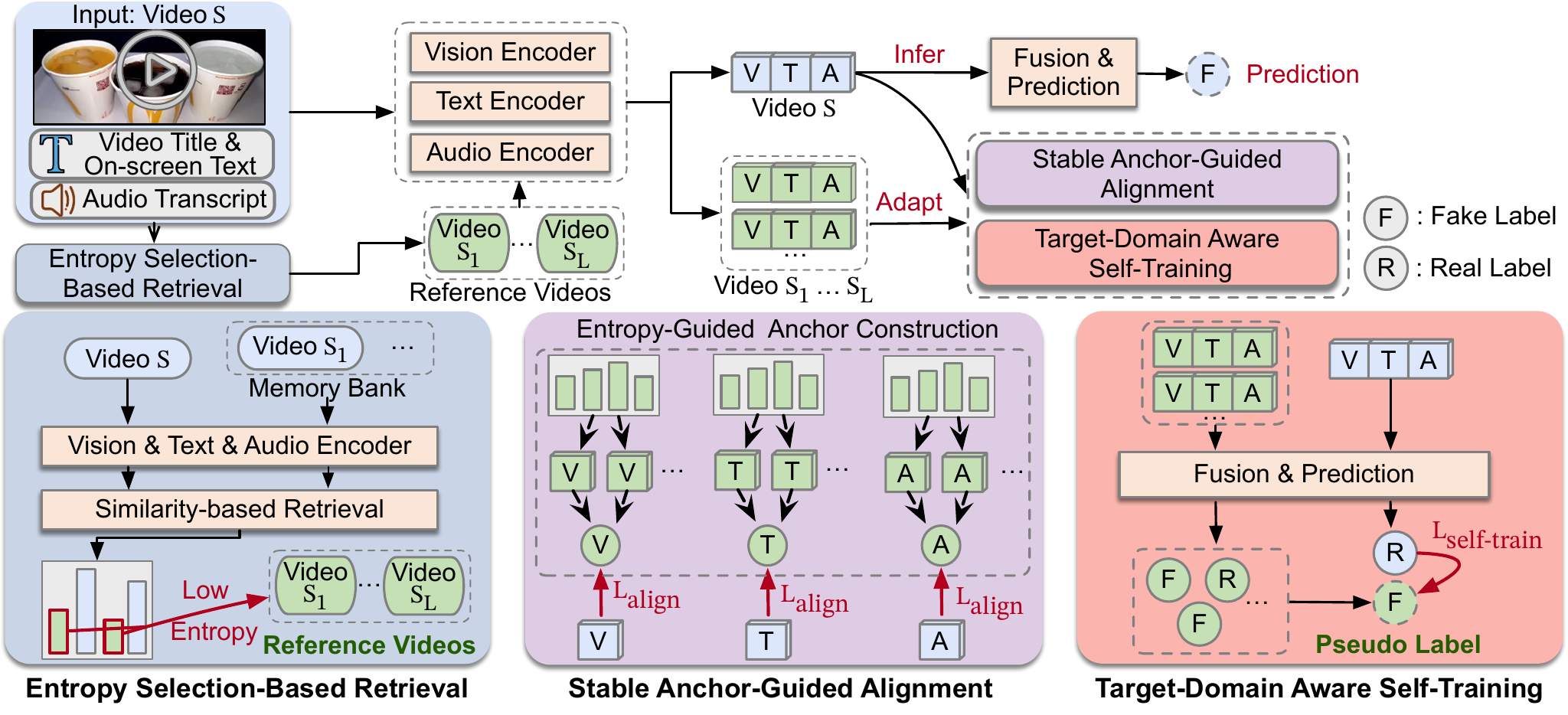}
  \vspace{-5mm}
  \caption{Overall framework of \M. (1) Entropy Selection-based Retrieval mechanism provides each target news video with semantically relevant yet stable reference instances.
  (2) Stable Anchor-Guided Alignment module explicitly narrows cross-domain gaps for the target news videos by matching their representations with their references.
  (3) Target-Domain Aware Self-Training paradigm endows news videos with reference-augmented category-aware pseudo-labels for self-training.
  }
  \label{fig:method}

\end{figure*}

\noindent \textbf{Our Pipeline.}
To cope with the unique yet significant challenges in achieving test-time \task, we introduce a novel framework dubbed \M that facilitates a retrieval-guided adaptation paradigm.
The overall framework of our method is illustrated in \Cref{fig:method}.
Specifically, \M first provides each news video $\mathcal{S}$ with $L$ semantically relevant and low-entropy reference instances: $\mathcal{R} = \{\mathcal{S}_i\}_{i=1}^{L}$ (\underline{\textbf{\Cref{sec:retrieval}}}).
Subsequently, \M explicitly aligns the representations of each news video with their low-entropy references at the distribution level (\underline{\textbf{\Cref{sec:alignment}}}).
Finally, \M endows each news video with a target-domain aware pseudo-label $\hat{y}_p$ with the guidance of their reference instances, and performs self-training using these pseudo-labels (\underline{\textbf{\Cref{sec:self-training}}}).

\subsection{Entropy Selection-Based Retrieval}
\label{sec:retrieval}
When testing a model trained on the source-domain dataset on a target domain for \task without adaptation, we empirically observe that not all news video instances associated with the same news events exhibit high uncertainty and prediction entropy (cf. \Cref{sec:intro} for illustrations). 
Target-domain news videos that share similar patterns with source video samples, such as similar themes and comparable deceptive editing techniques, tend to yield relatively low prediction-entropy when evaluated by the source model.
\begin{remark}
As has been highlighted in previous work~\cite{liang2020really, yuan2024tea, press2024entropy}, these low-entropy and stable target-domain news video instances have initial feature representations that are closer to the source domain and more accurate predictions by the source model.
\end{remark}
Based on these striking observations, we introduce a novel \textit{Entropy Selection-Based Retrieval mechanism}.
Under the source-free setting, it provides unstable target-domain news video instances with similar (probably tied to the same events) yet stable video instances in the target domain as references, which act as \textit{source-domain proxies} for assisting robust adaptation on the fly at test time.
Specifically, we maintain a memory bank $\mathcal{B}$ that continually stores the news videos in each newly arriving cluster: $\mathcal{B} = \{\mathcal{S}_i\}_{i=1}^{M}$.
To ensure efficiency and account for the fact that earlier news videos offer limited reference value for later ones, the memory bank operates in a First In, First Out (FIFO) manner and retains only the most recent $M$ news videos.
Subsequently, for each news video $\mathcal{S}$ as query, we retrieve its Top-$K$ semantically relevant video instances from the memory bank $\mathcal{B}$:
\begin{equation}\label{Eq:1}
\mathcal{R}' = \underset{\mathcal{S}_i \in \mathcal{B}}{\mathrm{Top\text{-}}K}(\text{Sim}(v, v_i) + \text{Sim}(t, t_i) + \text{Sim}(a, a_i)),
\end{equation}
where $\text{Sim}(\cdot)$ measures the intra-modal semantic relevance using cosine similarity, with vision modality as an example:
\begin{equation}\label{Eq:2}
    \text{Sim}(v, v_i) = \frac{{\Psi_v(v)}^{\top} \Psi_v(v_i)}{\|\Psi_v(v)\| \|\Psi_v(v_i)\|},
\end{equation}
here $\Psi_v(\cdot)$ is the off-the-shelf vision foundation model for multimodal retrieval (e.g., ViT~\cite{dosovitskiyimage}).
Based on the semantically relevant instances $\mathcal{R}' = \{\mathcal{S}_i\}_{i=1}^{K}$ for the query news video $\mathcal{S}$, we further filter out unstable reference videos via an entropy selection process:
\begin{equation}\label{Eq:3}
    \mathcal{R} =\{\mathcal{S}_i|\, \mathcal{S}_i \in \mathcal{R}', \text{Ent}_{\theta}(\mathcal{S}_i) < E_0\},
\end{equation}
where $\mathcal{R} = \{\mathcal{S}_i\}_{i=1}^{L}$ denotes a set of stable and semantically relevant reference instances for $\mathcal{S}$, $E_0$ is a predefined entropy threshold, and $\text{Ent}_{\theta}(\mathbf{x})$ represents the prediction entropy of sample $\mathbf{x}$:
\begin{equation}\label{Eq:4}
\text{Ent}_{\theta}(\mathbf{x}) = -\mathbf{p}_{\theta}(\mathbf{x}) \log \mathbf{p}_{\theta}(\mathbf{x}) 
= -\sum_{i=1}^{C} p_{\theta}^{(i)}(\mathbf{x}) \log p_{\theta}^{(i)}(\mathbf{x}),
\end{equation}
here $\mathbf{p}_{\theta}(\mathbf{x}) = \text{softmax}(\mathcal{F}_{\theta}(\mathbf{x})) = \{p_{\theta}^{(i)}(\mathbf{x})\}_{i=1}^{C}$ signifies the probabilistic distribution yielded by the source model $\mathcal{F}_{\theta}$, and $C$ denotes the number of classes ($C = 2$ for \task).
\textit{Notably}, not all news videos have available reference instances, since all semantically relevant candidates may be filtered out during the entropy selection step if they are deemed unstable, resulting in $\mathcal{R} = \emptyset$.
\textit{Nevertheless}, this occurrence rate is very low. 
And in the online setting, as the source model continually adapts to the target domain, more video instances stored in $\mathcal{B}$ tend to become stable references over time.

\subsection{Stable Anchor-Guided Alignment}
Mainstream TTA models commonly adopted Entropy Minimization (EM) based strategies to adapt the source model to the target domain by reducing the prediction entropy on unseen target-domain data~\cite{wang2021tent, niu2022efficient, guo2025smoothing}.
As a result, they improve the prediction confidence of the source model on the target samples, which \textit{implicitly} encourages the source model to yield source-like representations for these mild distribution-shifted (i.e., synthetic corruptions) instances~\cite{liang2020really, zhao2023pitfalls, press2024entropy}.
\textit{However}, these methods are not applicable to test-time \task with drastic topic-level distribution shifts:
 
\begin{remark}
Although EM-based methods implicitly align cross-domain representations, these indirect approaches struggle to handle the substantial discrepancy in news video representations caused by the topic-level distribution shifts in \task~\cite{zhao2023pitfalls} (cf.~\Cref{subsec:prelim}).
\end{remark}

\begin{algorithm}[t]
\caption{The pipeline of proposed \M.}
\label{alg:method}
\begin{algorithmic}[1]
\REQUIRE Target dataset $\mathcal{D}_{t} = \{\mathcal{S}_i\}_{i = 1}^{N_t} = \{(v_i, t_i, a_i)\}_{i = 1}^{N_t}$; source model $\mathcal{F}_{\theta} = \{f_{\theta_v}, f_{\theta_t}, f_{\theta_a}, \mathcal{P}_{\theta}\}$; memory bank $\mathcal{B}$; batch size $N_b$.
\ENSURE Final predictions $\{\hat{y}_i\}_{i=1}^{N_t}$ for all video samples in $\mathcal{D}_t$.

\FOR{each batch $\mathcal{X} = \{\mathcal{S}_i\}_{i=1}^{N_b}$ in $\mathcal{D}_t$}
    \STATE $/\ast$~\textit{\textbf{Entropy Selection-Based Retrieval}}~$\ast/$
    \STATE Update memory bank $\mathcal{B}$ for $\mathcal{S} \in \mathcal{X}$ in a FIFO manner.
    \STATE Retrieve semantically relevant reference news video set $\mathcal{R}'$ for all $\mathcal{S} \in \mathcal{X}$ using Eq.~(\ref{Eq:1}) -- (\ref{Eq:2}).
    \STATE Filter unstable references to obtain stable (low-entropy) reference set $\mathcal{R}$ for all $\mathcal{S} \in \mathcal{X}$ using Eq.~(\ref{Eq:3}) -- (\ref{Eq:4}).
     \STATE $/\ast$~\textit{\textbf{Stable Anchor-Guided Alignment}}~$\ast/$
    \STATE Compute modality-specific stable anchor representation $\mathbf{A}_m(\mathcal{S})$, using Eq.~(\ref{Eq:5}).
    \STATE Calculate alignment loss $\mathcal{L}_{\text{align}}$ via Eq.~(\ref{Eq:6})).
    \STATE $/\ast$~\textit{\textbf{Target-Domain Aware Self-Training}}~$\ast/$
    \STATE Obtain prediction probability $\hat{p}$ for all $\mathcal{S} \in \mathcal{X}$ and $\hat{p}_i$ for each of their reference video $\mathcal{S}_i \in \mathcal{R}$.
    \STATE Compute target-domain aware pseudo-label $\hat{y}_p$ for all $\mathcal{S} \in \mathcal{X}$ and self-training loss $\mathcal{L}_{\text{self-train}}$ using Eq.~(\ref{Eq:11}) -- (\ref{Eq:12}).
    \STATE Update parameters of $\mathcal{F}_{\theta}$ with Eq.~(\ref{Eq:13}).
    \STATE Yield final predictions for $\mathcal{S} \in \mathcal{X}$ with updated $\mathcal{F}_{\theta}$.
\ENDFOR
\end{algorithmic}
\end{algorithm}

In contrast, we propose a new \textit{Stable Anchor-Guided Alignment module}, which \textit{explicitly} bridges the cross-domain representation gap by only using the unlabeled target data themselves.
Specifically, for a news video $\mathcal{S}$, 
we establish modality-specific stable anchors by applying an entropy-guided construction mechanism to its retrieved source-proxy news videos
\begin{equation}
\label{Eq:5}
    \mathbf{A}_{m}(\mathcal{S}) = \sum_{i=1}^{L} \frac{\exp(-\text{Ent}_{\theta}(\mathcal{S}_i))}{\sum_{j=1}^{L} \exp(-\text{Ent}_{\theta}(\mathcal{S}_j))} f_{{\theta}_{m}}(m_i), m \in \{v,t,a\}
\end{equation}
where $f_{{\theta}_{m}}(m_i) \in\mathbb R^{D_m}$ is the representation for modality $m$ for reference video $\mathcal{S}_i$, and $D_m$ is feature dimension.

\begin{remark}
The stable anchor $\mathbf{A}_{m}(\mathcal{S})$, constructed from entropy-weighted representations from low-entropy news videos, is stable and closer to the source-domain distribution.
Moreover, this anchor representation shares similar semantics with the query news video, ensuring that the alignment is both meaningful and stable.
\end{remark}

Building upon the ``stable anchor'', we explicitly align the representation of the news video $\mathcal{S}$ with the anchor representation through the prototype-based alignment~\cite{zhang2023properties}, drawing its representation closer to the source domain:
\begin{equation}\label{Eq:6}
    \mathcal{L}_{\text{align}} = \sum_{m \in \mathcal{M}} \left( 1 - \frac{f_{{\theta}_{m}}(m)^\top \mathbf{A}_m(\mathcal{S})}{\| f_{{\theta}_{m}}(m) \| \| \mathbf{A}_m(\mathcal{S}) \|} \right)
\end{equation}

\label{sec:alignment}

\subsection{Target-Domain Aware Self-Training}
\label{sec:self-training}
The varying popularity and public attention of different news events contribute to a severely imbalanced, ever-changing, and unpredictable category distribution (real \textit{vs.} fake) in newly arriving video clusters associated with diverse budding events~\cite{qi2023fakesv, bu2024fakingrecipe, papadopoulou2019corpus}.
\textit{However}, prior works have demonstrated that most TTA methods perform well only when the target domain's label distribution remains stable and consistent with that of the source domain~\cite{zhao2023pitfalls}.
To address this challenge, a natural solution in conventional settings is to fine-tune the model on each new distribution to facilitate the adaptation to the imbalance in label proportions.
Motivated by this, we propose a novel \textit{Target-Domain Aware Self-Training paradigm} under the TTA setting, which allows the model to adapt to the imbalanced and ever-changing category distribution on the fly at test time, without requiring golden labels.
Specifically, we first endow each news video instance $\mathcal{S}$ with pseudo-label $\hat{y}_p$ under the guidance of its stable and semantically similar reference videos:
\begin{equation}
\label{Eq:11}
\hat{y}_p = \arg\max\limits_{c}\,\Big( \alpha \cdot \hat{p}^c + \beta \cdot  \sum_{i=1}^{L} \frac{\exp\!\big(\operatorname{Sim}(\mathcal{S}, \mathcal{S}_i)\big)}
{\sum_{j=1}^{L} \exp\!\big(\operatorname{Sim}(\mathcal{S}, \mathcal{S}_j)\big)}\, \hat{p}^c_i \Big), 
\end{equation}
where $\hat{p}^c$ and $\hat{p}_i^c$ are prediction probabilities of class $c$ for the video $\mathcal{S}$ and its reference $\mathcal{S}_i$ from source model $\mathcal{F}_{\theta}$, respectively, $\alpha$ and $\beta$ are the weighting factors, and $\operatorname{Sim}(\mathcal{S},\mathcal{S}_i)$ is the sum of similarity scores across all modalities between $\mathcal{S}$ and $\mathcal{S}_i$ (refer to Eq.~(\ref{Eq:1})).

\begin{remark}
The target-domain aware pseudo-label $\hat{y}_p$ of the news video $\mathcal{S}$ is derived with the augmentation of the pseudo-labels from its stable (low-entropy) and semantically relevant reference videos.
As a result, it is accurate and can well reflect the category distribution of the news event cluster to which the query news video belongs.
\end{remark}

\label{sec:experiments}
\begin{table*}[t]
\centering
\caption{Performance comparison under random batch sampling. The best results are in black bold while the second are underlined. Higher values of Accuracy (Acc), Macro-F1 (M-F1), and Macro-Recall (M-R) indicate better performance.}
\vspace{-2mm}
\label{tab:main_result_random}
\setlength{\tabcolsep}{2.5pt}
\renewcommand{\arraystretch}{1.2}
\resizebox{\linewidth}{!}{
\begin{tabular}{l|ccc|ccc|ccc|ccc|ccc|ccc}
\toprule
 &   \multicolumn{3}{c|}{\textbf{FakeTT \textrightarrow~FakeSV}} & \multicolumn{3}{c|}{\textbf{FakeTT \textrightarrow~FVC}} & \multicolumn{3}{c|}{\textbf{FakeSV\textrightarrow~FakeTT}} & \multicolumn{3}{c|}{\textbf{FakeSV \textrightarrow~FVC}} & \multicolumn{3}{c|}{\textbf{FVC \textrightarrow~FakeSV}} & \multicolumn{3}{c}{\textbf{FVC\textrightarrow~FakeTT}} \\ 
    \cmidrule(lr){2-4} \cmidrule(lr){5-7} \cmidrule(lr){8-10} \cmidrule(lr){11-13} \cmidrule(lr){14-16} \cmidrule(lr){17-19}
\textbf{Model} & \textbf{Acc} & \textbf{M-F1} & \textbf{M-R} & \textbf{Acc} & \textbf{M-F1} & \textbf{M-R} & \textbf{Acc} & \textbf{M-F1} & \textbf{M-R} & \textbf{Acc} & \textbf{M-F1} & \textbf{M-R} & \textbf{Acc} & \textbf{M-F1} & \textbf{M-R} & \textbf{Acc} & \textbf{M-F1} & \textbf{M-R}\\
\midrule
FANVM & 51.43 & 42.21 & 51.39 & 55.03 & 55.01 & 57.27 & 57.48 & 45.13 & 50.65 & 53.26 & 44.80 & 47.72 & 50.91 & 49.35 & 50.89 & 47.69 & 46.88 & 46.98 \\
SV-FEND & 53.53 & 50.36 & 53.56 & 58.50 & 56.96 & 56.94 & \underline{63.00} & 60.06 & 60.07 & 61.00 & 45.02 & 52.95 & 57.34 & 57.07 & 57.33 & 56.33 & 56.32 & 58.09 \\
FakRec & 62.50 & 62.24 & 62.49 & 60.24 & 58.83 & 58.82 & 61.09 & 55.74 & 60.45 & 63.02 & 52.85 & 56.45 & 60.18 & 59.10 & 60.16 & 61.95 & 61.87 & 63.16 \\

ExMRD & 50.52 & 34.77 & 50.47 & 61.25 & 58.02 & 59.13 & 61.85 & 46.70 & 54.08 & 63.42 & 51.76 & 56.35 & 50.19 & 33.81 & 50.14 & 56.88 & 56.71 & 60.54 \\
REAL & 54.75 & 51.19 & 54.78 & 58.61 & 52.12 & 53.69 & 60.04 & 54.52 & 55.60 & 60.96 & 45.54 & 53.05 & 57.59 & 57.12 & 57.58 & 60.19 & 59.64 & 60.01 \\
\midrule
Source & 62.47 & 59.77 & 62.44 & 56.55 & 55.94 & 56.27 & 59.69 & 40.21 & 51.16 & \underline{63.93} & 49.57 & 56.09 & 59.00 & 53.68 & 58.96 & 61.85 & 61.85 & 63.86 \\
Tent & 64.74 & 64.49 & 64.74 & 60.53 & 56.65 & 56.97 & 60.49 & 46.34 & 53.04 & 61.40 & 46.31 & 53.54 & 63.49 & 61.81 & 63.52 & 61.70 & 61.69 & 63.55 \\
SAR & \underline{65.56} & \underline{64.95} & \underline{65.58} & \underline{61.83} & \underline{59.57} & \underline{59.50} & 60.04 & 45.53 & 52.53 & 61.29 & 45.96 & 53.38 & 64.46 & \underline{63.64} & 64.48 & 63.05 & 63.02 & \underline{64.59} \\
READ & 64.02 & 63.83 & 64.03 & 56.40 & 56.04 & 56.61 & 62.90 & \underline{62.55} & \underline{63.17} & 63.89 & \underline{63.36} & \underline{63.80} & \underline{64.62} & 63.52 & \underline{64.64} & 63.40 & \underline{63.25} & 64.28 \\

SuMi & 58.83 & 55.93 & 58.86 & 60.60 & 56.23 & 56.74 & 59.09 & 59.09 & 60.99 & 60.96 & 60.90 & 62.27 & 55.38 & 47.54 & 55.42 & \underline{63.55} & 57.72 & 58.81 \\
\midrule
\rowcolor{gray!12}

\textbf{\M} & \textbf{67.61} & \textbf{67.60} & \textbf{67.61} & \textbf{63.44} & \textbf{62.83} & \textbf{63.21} & \textbf{64.71} & \textbf{63.49} & \textbf{63.46} & \textbf{66.68} & \textbf{65.05} & \textbf{64.82} & \textbf{66.87} & \textbf{66.36} & \textbf{66.89} & \textbf{65.52} & \textbf{64.70} & \textbf{64.86} \\

\bottomrule
\end{tabular}
}
\vspace{-2mm}
\end{table*}

Subsequently, we define the self-training objective function as: 
\begin{equation}\label{Eq:12}
    \mathcal{L}_{\text{self-train}} = - \sum_{c=1}^{C} \hat{y}_p^c \log \hat{p}^c,
\end{equation}
where $C$ is the total number of classes. At test time, the source model $\mathcal{F}_{\theta}$ is optimized with the following loss function:
\begin{equation}\label{Eq:13}
        \mathcal{L}_{\text{total}} =  \gamma \mathcal L_{\text{align}} + \mathcal{L}_{\text{self-train}} + \mathcal{L}_{\text{entropy}},
\end{equation}
where $\mathcal{L}_{\text{entropy}}$ is the widely adopted EM objective in TTA~\cite{wang2021tent, liang2020really, yang2024testtime, guo2025smoothing} which minimizes the self-entropy of prediction (refer to~Eq.(\ref{Eq:4})), and $\gamma$ is loss balancing factor for alignment loss.
In line with prior works~\cite{wang2021tent,liang2020really, yang2024testtime}, only the last layer from modality-specific encoders and normalization layers are learnable to ensure efficient and stable adaptation.
We adopt the most applicable online TTA setting~\cite{wang2021tent, yang2024testtime, guo2025smoothing}, where the source model $\mathcal{F}_{\theta}$ is first optimized with the current batch of news videos, and then yields the final prediction for this batch on the fly.
\textit{Notably}, although our primary goal is to enable the source model to effectively adapt to unstable and high-entropy news video instances, we do not exclude the optimization of low-entropy instances, as they contribute to stable and reliable training~\cite{niu2022efficient, niu2023towards}.
The pipeline of our proposed \M is summarized in \Cref{alg:method}.

\section{Experiments}

\begin{table}[t]

  \centering
  \setlength{\tabcolsep}{2pt}
  \caption{Statistics of datasets. \# denotes the number of items.}
  \vspace{-3mm}
  \label{tab:dataset}
  \resizebox{\linewidth}{!}{
  \begin{tabular}{@{}ccccccc@{}}
    \toprule
    Dataset &Time Range &\#~Fake &\#~Real &\#~Event & Language & Platform\\

    \midrule
    FakeTT    &2019/05-2024/03  &1,172 &819 & 286 & English & TikTok \\
    FakeSV	&2017/10-2022/02    &1,810 &1,814 & 738 & Chinese & Douyin \\
    FVC & 2016/01-2018/01 & 1,633  & 1,131 & 305  & English  & YouTube \\
    \bottomrule
  \end{tabular}
  }

\end{table}

\subsection{Experimental Setup}

In this section, we provide an overview of the experimental setup, with a detailed version available at \Cref{app:setup}.

\begin{figure}[t]

    \centering
    \includegraphics[width=\columnwidth]{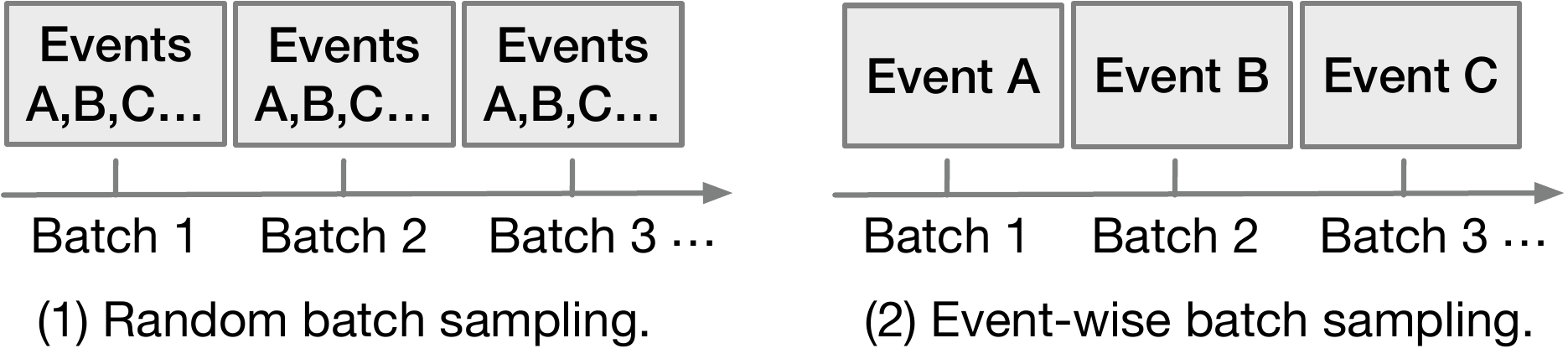}
    \vspace{-6mm}
    \caption{Random \textit{vs.} Event-wise batch sampling.}
    \label{fig:sampling}
    \vspace{-3mm}
\end{figure}

\noindent \textbf{TTA Setting.} 
To simulate significant topic-level shifts in the evolution of fake news videos, we select three public datasets for \task that are mutually disjoint in event time ranges or differ in language and platform.
As a result, any two datasets have no overlapping events and exhibit a drastic covariate shift. 
We set one dataset as the source domain and the other two as target domains, resulting in six evaluation groups for test-time \task (e.g., FakeTT \textrightarrow~FakeSV).
For each evaluation group, the source model is first trained on the source-domain dataset and then performs test-time adaptation on the unlabeled target domain using different adaptation methods (i.e., TTA baselines and our \M).
We adopt the most popular \textit{online adaptation} setting that the evaluation happens immediately after the arrival of each batch of data~\cite{wang2021tent, niu2022efficient, guo2025smoothing}.
Consequently, as shown in~\Cref{fig:sampling}, we define two batch sampling types for evaluation: (1) \textit{random batch sampling}, where each batch is randomly drawn from the target domain, and (2) \textit{event-wise batch sampling}, which reduces the batch size to approximate the typical event size and strives to make each batch represent a single incoming event. 
The event-wise batch sampling setting is more realistic and challenging, since the topics shift frequently and the label distributions are unpredictable, highly imbalanced, and varying across batches.

\noindent \textbf{Datasets.}
We adopt three datasets in \task that differ in either event time ranges or platform and language: 
FakeTT~\cite{bu2024fakingrecipe}, FakeSV~\cite{qi2023fakesv}, and
FVC~\cite{papadopoulou2019corpus},
with detailed statistics in~\Cref{tab:dataset}.

\noindent \textbf{Baselines.}
We compare our proposed \M with \NumBaseline competitive baselines, which can be divided into two groups:
(1) \textit{Traditional \task methods} that are trained on the source dataset without adaptation, including FANVM~\cite{choi2021using}, SV-FEND~\cite{qi2023fakesv}, FakRec~\cite{bu2024fakingrecipe}, ExMRD~\cite{hong2025following}, and REAL~\cite{li2025real};
(2) \textit{TTA methods}, including Source Model (Source), Tent~\cite{wang2021tent}, SAR~\cite{niu2023towards}, READ~\cite{yang2024testtime}, and SuMi~\cite{guo2025smoothing}.

\noindent \textbf{Implementation Details.}
The source model is implemented with feedforward networks (FFN) as modality-specific encoders, a two-layer transformer as the modal fusion module, and an MLP as the classifier.
We adopt video frames as visual modality information, video titles and on-screen text as textual modality information, and audio transcripts as audio modality.
We adopt pre-trained ViT~\cite{dosovitskiyimage} and BERT~\cite{devlin2019bert} for feature extraction and multimodal retrieval (More details are present in~\Cref{app:feature}).
The adaptation process uses an initial learning rate of 0.0001 within a single epoch.

\begin{table*}[t]
\centering
\caption{Performance comparison under event-wise batch sampling. The best results are in black bold. Higher values of Accuracy (Acc), Macro-F1 (M-F1), and Macro-Recall (M-R) indicate better performance.}
\vspace{-2mm}
\label{tab:main_result_event}
\setlength{\tabcolsep}{2.5pt}
\renewcommand{\arraystretch}{1.2}
\resizebox{\linewidth}{!}{
\begin{tabular}{l|ccc|ccc|ccc|ccc|ccc|ccc}
\toprule
 &  \multicolumn{3}{c|}{\textbf{FakeTT\textrightarrow~FakeSV}} & \multicolumn{3}{c|}{\textbf{FakeTT \textrightarrow~FVC}} & 
 \multicolumn{3}{c|}{\textbf{FakeSV \textrightarrow~FakeTT}} & \multicolumn{3}{c|}{\textbf{FakeSV \textrightarrow~FVC}} & \multicolumn{3}{c|}{\textbf{FVC \textrightarrow~FakeSV}} & \multicolumn{3}{c}{\textbf{FVC\textrightarrow~FakeTT}} \\ 
    \cmidrule(lr){2-4} \cmidrule(lr){5-7} \cmidrule(lr){8-10} \cmidrule(lr){11-13} \cmidrule(lr){14-16} \cmidrule(lr){17-19}
\textbf{Model} & \textbf{Acc} & \textbf{M-F1} & \textbf{M-R} & \textbf{Acc} & \textbf{M-F1} & \textbf{M-R} & \textbf{Acc} & \textbf{M-F1} & \textbf{M-R} & \textbf{Acc} & \textbf{M-F1} & \textbf{M-R} & \textbf{Acc} & \textbf{M-F1} & \textbf{M-R} & \textbf{Acc} & \textbf{M-F1} & \textbf{M-R}\\
\midrule

Source & 62.47 & 59.77 & 62.44 & 56.53 & 55.93 & 56.26 & 59.81 & 40.27 & 51.17 & \underline{63.92} & 49.56 & 56.09 & 59.00 & 53.68 & 58.96 & 61.82 & \underline{61.82} & \textbf{63.86} \\
Tent & 52.35 & 42.54 & 52.39 & \underline{59.10} & 40.29 & 50.59 & 59.15 & 37.74 & 50.22 & 59.17 & 37.75 & 50.19 & 51.99 & 38.92 & 52.04 & 59.51 & 39.82 & 50.86 \\

SAR & 50.94 & 37.63 & 50.99 & 58.92 & 39.70 & 50.34 & 59.10 & 38.05 & 50.24 & 41.19 & 29.77 & 50.15 & 50.41 & 34.67 & 50.47 & 59.41 & 39.36 & 50.70 \\
READ & \underline{64.32} & \underline{64.26} & \underline{64.32} & 57.51 & \underline{57.04} & \underline{57.50} & \underline{61.27} & \underline{60.89} & \textbf{61.49} & 62.87 & \underline{62.48} & \underline{63.10} & \underline{64.46} & \underline{64.46} & \underline{64.46} & \underline{62.88} & 54.43 & 56.97 \\

SuMi & 56.93 & 56.82 & 56.93 & 55.70 & 54.36 & 54.38 & 58.30 & 57.82 & 58.26 & 56.61 & 56.39 & 57.21 & 60.02 & 59.59 & 60.03 & 59.81 & 55.52 & 56.04 \\
\midrule
\rowcolor{gray!12}

\textbf{\M} & \textbf{65.76} & \textbf{65.72} & \textbf{65.75} & \textbf{60.82} & \textbf{60.13} & \textbf{60.43} & \textbf{63.08} & \textbf{61.41} & \underline{61.30} & \textbf{65.17} & \textbf{63.35} & \textbf{63.15} & \textbf{65.45} & \textbf{65.01} & \textbf{66.31} & \textbf{64.59} & \textbf{63.23} & \underline{63.16}\\ 

\bottomrule
\end{tabular}
}

\end{table*}

\begin{figure}[t]
  \centering
  \includegraphics[width=0.9\columnwidth]{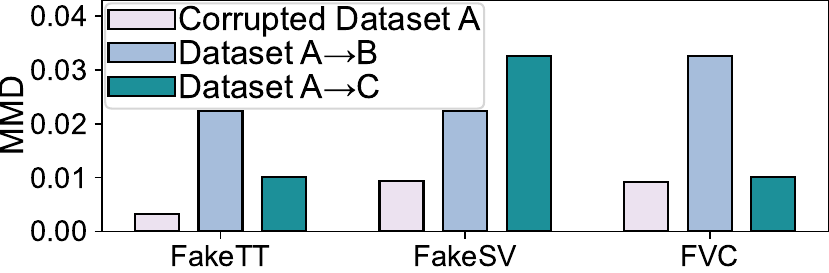}
  \vspace{-3.5mm}
  \caption{Severity of distribution shift caused by data corruptions in prior TTA methods \textit{vs}. topic-level shifts in \task.}
  \vspace{-3mm}
  \label{fig:exp-pre}
\end{figure}

\subsection{Preliminary Experiment}
\label{subsec:prelim}
We empirically compare the severity of distribution shift between the prior TTA works and test-time \task in~\Cref{fig:exp-pre}.
Specifically, we first sample 1,000 videos from the FakeTT, FakeSV, and FVC datasets.
For each dataset, following TTA studies~\cite{hendrycks2019benchmarking, wang2021tent, yang2024testtime}, we add Gaussian noise corruptions at the highest severity level (level 5) to the clean video frames and randomly apply two types of textual corruptions adopted from prior work~\cite{zhang2024benchmarking} for clean text and audio transcripts.
We compute the Maximum Mean Discrepancy (MMD) distance~\cite{gretton2006kernel, jayasumana2024rethinking} with a Gaussian RBF kernel between each corrupted dataset (on feature level) and its clean counterpart, as well as the MMD distance between the clean version of that dataset and the clean versions of the other two datasets (the detailed formula is in~\Cref{app:mmd}).
The latter corresponds to the scenario adopted in our experiments to simulate real-world fake news video evolution.
The final MMD distance is calculated by summing over the single modality distances.
As illustrated in~\Cref{fig:exp-pre}, the distribution shift caused by topic-level fake news video evolution is \textit{more severe} than the data corruptions encountered by prior TTA methods.
Therefore a more robust and effective solution is needed for test-time \task.

\begin{table}[t]
    \centering
    \caption{Ablation study on core components in \M under random batch sampling. FSV: FakeSV, FTT: FakeTT.}
    \vspace*{-2mm}
    \label{tab:ablation}
        \setlength{\tabcolsep}{3.5pt}
    \resizebox{\linewidth}{!}{
    \begin{tabular}{cccccc}
        \toprule
         & & \multicolumn{2}{c}{\textbf{FTT \textrightarrow~FSV}} & \multicolumn{2}{c}{\textbf{FTT \textrightarrow~FVC}}\\
         
         \cmidrule(lr){3-4} \cmidrule(lr){5-6} 
         
       \textbf{Module} &  \textbf{Variant} & \textbf{Acc} & \textbf{M-F1} &\textbf{Acc} & \textbf{M-F1} \\
        \midrule
     \multirow{2}{*}{\makecell[c]{Retrieval}} 
    & w/o Similarity Retrieval  & 63.52 & 63.37 & 62.71 & 61.92 \\ 
    & w/o Entropy Selection   & 61.98 & 61.98   & 58.01 & 57.15  \\
    \midrule
    \multirow{2}{*}{\makecell[c]{Alignment}} 
    & w/ MSE Alignment         & 62.53 & 62.27 & 63.04 & 62.38 \\ 
    & w/o Alignment    & 60.49 & 60.47 & 62.35 & 61.62  \\
    \midrule
    \multirow{2}{*}{\makecell[c]{Self-\\Training}} 
    & w/ Self-Labeling  & 64.57 & 64.24 & 50.01 & 49.84 \\ 
    & w/o Self-Training  & 65.62 & 65.58 & 57.83 & 56.99 \\
    \midrule
    \rowcolor{gray!12}
    \textbf{\M} & \textbf{ALL}   & \textbf{67.61} & \textbf{67.60} & \textbf{63.44} & \textbf{62.83}  \\ 
     \bottomrule
    \end{tabular}
    }
    \vspace{-3mm}
\end{table}

\subsection{Main Performance}
\label{subsec:main}
\subsubsection{Performance under Random Batch Sampling}
We first compare \M with all \NumBaseline baselines across all six evaluation groups under random batch sampling setting, with results reported in~\Cref{tab:main_result_random}. 
From the results, we have the following observations:

\textbf{(O1)} \textbf{Traditional \task methods} that assume an identical training and test data distribution fail to tackle the evolution of fake news videos with drastic distribution shift.
\textbf{TTA methods} present a certain degree of effectiveness in adapting to the distribution-shifted target domain, with relatively consistent improvements in performance compared to the source model.
\textit{However}, these methods that adopted EM-based strategies still struggle to tackle the drastic news video representation gaps caused by topic-level distribution shift in \task.
\textbf{(O2)} \textbf{\M} significantly outperforms all competitive baselines with an average improvement of 6.55\% in terms of Macro-F1 across all evaluations. 
This substantial performance gain stems from our novel retrieval-guided adaptation paradigm. 
Specifically, this paradigm explicitly aligns the representations of unstable news video instances with their semantically similar and stable reference videos, effectively bridging cross-domain gaps. 
In addition, the self-training paradigm leverages highly informative pseudo-labels derived from augmentations of stable reference videos, which further enhances robust test-time adaptation for the \task.

\subsubsection{Performance under Event-Wise Batch Sampling}
We further compare \M with 6 TTA baselines under a more challenging event-wise batch sampling setting that better reflects real-world news video evolution, with results in~\Cref{tab:main_result_event}.
Under this complex setting, \textit{where topic-level distribution shifts occur across batches and label distributions are severely imbalanced and vary across batches}, baselines designed for mild shift scenarios, such as Tent~\cite{wang2021tent} and READ~\cite{yang2024testtime}, even fail to adapt to the target domain.
SAR~\cite{niu2023towards} alleviates the challenges by combining batch-agnostic normalization optimization with parameter recovery.
SuMi~\cite{guo2025smoothing} handles this complex scenario to some extent through a smoothing adaptation strategy.
\textit{Nevertheless}, \textbf{\M} can still outperform all these baselines with a large margin, achieving an average Macro-F1 improvement of 5.45\%.
The Stable Anchor-Guided Alignment module enables \M to handle drastic and rapidly changing news video representations under such conditions.
Moreover, the Target-Domain Aware Self-Training paradigm provides each incoming news video with a target-domain aware pseudo-label that accurately captures the category distribution of the associated news event cluster,
assisting \M to promptly adapt to the current label distribution. 

\subsection{Ablation Study}
To explore the role of each component in \M, we conduct comprehensive ablations and the results under \textit{random batch sampling setting} are in~\Cref{tab:ablation}. 
To validate the efficacy of \textit{Entropy Selection-Based Retrieval mechanism}, we design two variants:
\textbf{(1) w/o Similarity Retrieval}, and \textbf{(2) w/o Entropy Selection}, where reference instances for each news video are selected based solely on entropy or similarity, respectively.
Both variants suffer significant performance degradation, as aligning representations to irrelevant or unstable (high-entropy) reference videos leads to unreliable adaptation and potential model collapse.
To assess the efficacy of \textit{Stable Anchor-Guided Alignment module}, we introduce two variants: \textbf{(1) w/ MSE Alignment}, which replaces original alignment with a L2-norm minimization between the features of the query news video and the averaged features from its reference instances, and \textbf{(2) w/o Alignment}, where the alignment module is entirely removed.
The MSE alignment results in a drastic performance drop, as it performs simple point-to-point alignment, which ignores the underlying semantic structure of the representations and enforces rigid feature matching, resulting in distorted features.
Removing the alignment module also leads to suboptimal performance, as the drastic cross-domain representation discrepancies caused by topic-level shifts in \task necessitate a more explicit remedy.
To evaluate the efficacy of \textit{Target-Domain Aware Self-Training paradigm}, we propose two variants: \textbf{(1) w/ Self-Labeling}, where the pseudo-label is generated from the query news video without the augmentation from its references, and \textbf{(2) w/o Self-Training}, where the self-training is eliminated.
Both variants lead to performance drop, as the pseudo-labels derived under the guidance of stable reference news videos are more informative and reliable.

\begin{figure}[t]
    \centering
    \begin{subfigure}[b]{0.50\columnwidth}
    \centering
    \includegraphics[width=\columnwidth]{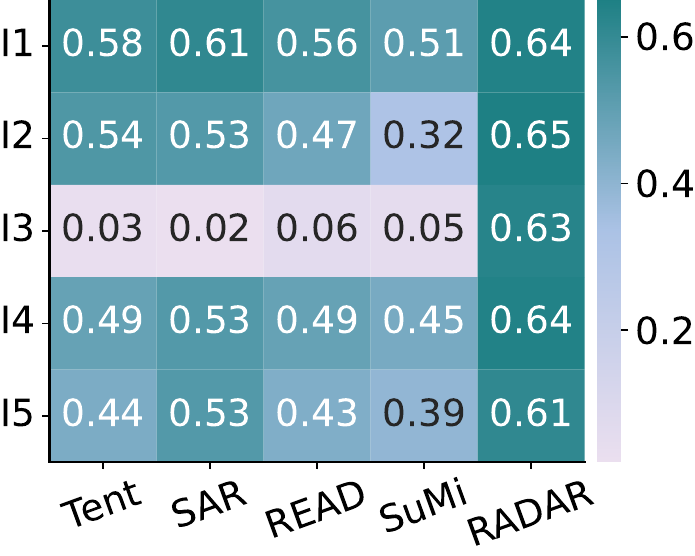}
    \vspace{-3mm}
    \caption{From FakeTT to FakeSV.}
    \end{subfigure}
    \hfill
    \begin{subfigure}[b]{0.49\columnwidth}
    \centering
    \includegraphics[width=\columnwidth]{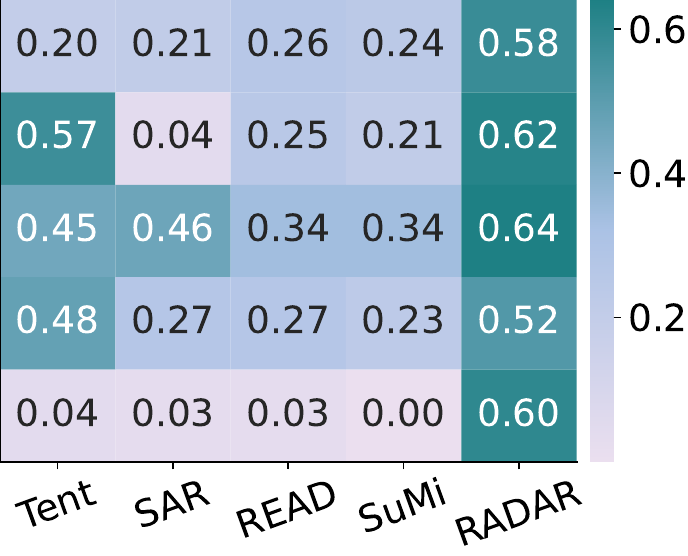}
    \vspace{-3mm}
    \caption{From FakeTT to FVC.}
    \end{subfigure}
    \vspace{-6mm}
    \caption{Entropy reduction (larger is better) on five unstable news videos among \M and baselines. I: Instance.}
    \label{fig:entropy_variations}

\end{figure}

\begin{figure}[t]
    \centering
    \begin{subfigure}[b]{0.99\columnwidth}
    \centering
    \includegraphics[width=\columnwidth]{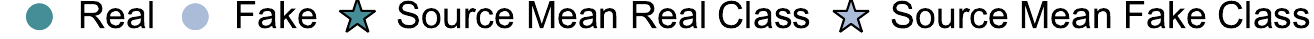}
    \end{subfigure}
    \centering
    \begin{subfigure}[b]{0.49\columnwidth}
    \centering
    \includegraphics[width=\columnwidth]{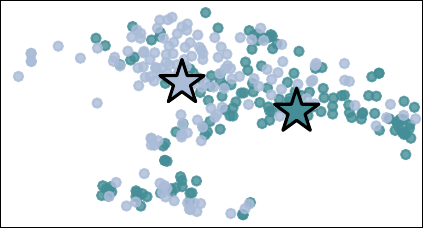}
    \caption{Baseline Tent.}
    \end{subfigure}
    \hfill
    \begin{subfigure}[b]{0.49\columnwidth}
    \centering
    \includegraphics[width=\columnwidth]{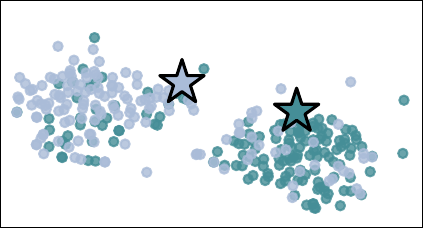}
    \caption{Our proposed \M.}
    \end{subfigure}
    \vspace{-3mm}
    \caption{Visualization of cross-domain representations.}
    \label{fig:visualization}
    \vspace{-3mm}
\end{figure}

\subsection{Quantitative Analysis of Entropy Variation}
To evaluate the effectiveness of proposed \M in adapting the source model to severely unstable news videos, we randomly select five instances with high initial prediction entropy from the FakeSV and FVC datasets, respectively.
We track their entropy reductions and compare the performance of \M against four competitive TTA baselines under \textit{random batch sampling setting}.
As illustrated in~\Cref{fig:entropy_variations}, \M significantly reduces the entropy of these unstable instances compared to the baselines, owing to its novel \textit{retrieval-guided adaptation paradigm}.
By leveraging stable reference news videos, \M explicitly aligns the representations of high-entropy video samples toward the source domain, enabling more confident and robust prediction.

\subsection{Visualization of Representation Alignment}
\label{subsec:exp-visualization}
To evaluate \M in cross-domain representation alignment, we visualize the news video representations produced by our proposed \M and the baseline Tent~\cite{wang2021tent} on the target FakeSV dataset using FakeTT as the source, under the \textit{random batch sampling} setting.
Following~\cite{press2024entropy}, we jointly visualize the class-wise (real and fake) mean embeddings produced by the source model on the source data, and the feature embeddings of 500 selected target videos produced by \M and Tent, using T-SNE~\cite{van2008visualizing}.
The embeddings are the outputs of the modal-specific encoders, and the three modalities are averaged to obtain the final representation.
As shown in~\Cref{fig:visualization}, the target-domain representations derived by \M are more clustered and closely aligned with the mean embeddings of corresponding classes from the source data, highlighting the efficacy of \textit{Stable Anchor-Guided Alignment module} in bridging the cross-domain representation gap.

\subsection{Evaluation of Pseudo-Label Accuracy}
\begin{figure}[t]
    \centering
    \begin{subfigure}[b]{0.5\columnwidth}
    \centering
    \includegraphics[width=\columnwidth]{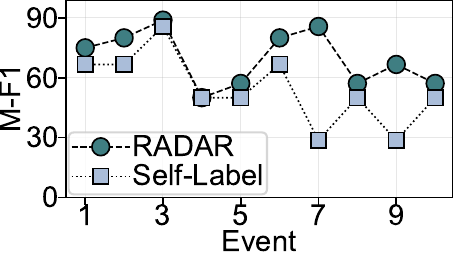}
    \vspace{-5mm}
    \caption{From FakeTT to FakeSV.}
    \end{subfigure}
    \hfill
    \begin{subfigure}[b]{0.49\columnwidth}
    \centering
    \includegraphics[width=\columnwidth]{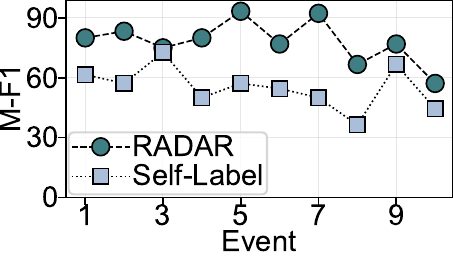}
    \vspace{-5mm}
    \caption{From FakeTT to FVC.}
    \end{subfigure}
    \vspace{-8mm}
    \caption{Pseudo-label accuracy between \M and self-labeling method on category-imbalanced events.}
    \label{fig:pseudo_label}
    \vspace{-4mm}
\end{figure}

To assess the effectiveness of \M in handling highly imbalanced and dynamically varying category distributions within news video clusters, we compare the pseudo-label accuracy generated by \M with that of a standard self-labeling approach.
This evaluation is conducted under the \textit{event-wise batch sampling setting}, where the category distributions are severely unbalanced and fluctuate significantly across batches.
Specifically, we randomly select 10 events from the target datasets FakeSV and FVC, focusing on those in which the associated news video instances exhibit a label imbalance ratio greater than 8:1.
For each event, we calculate the M-F1 scores of the pseudo-labels for the news videos tied to that event.
As illustrated in~\Cref{fig:pseudo_label}, \M produces substantially more accurate pseudo-labels than the baseline, showcasing the effectiveness of the proposed \textit{Target-Domain Aware Self-Training paradigm} in generating robust and informative supervision signals under unpredictable and imbalanced label distributions.

\begin{figure*}[t]
    \centering
    \begin{subfigure}[b]{0.31\textwidth}
        \centering
        \includegraphics[width=\linewidth]{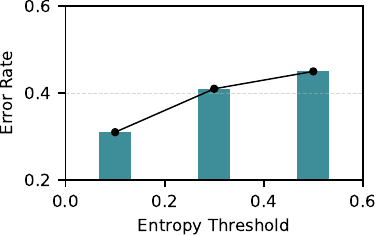}
        \vspace{-3mm}
        \caption{From FakeTT to FakeSV.}
    \end{subfigure}
    \hfill
    \begin{subfigure}[b]{0.31\textwidth}
        \centering
        \includegraphics[width=\linewidth]{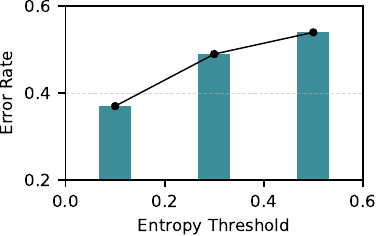}
        \vspace{-3mm}
        \caption{From FakeTT to FVC.}
    \end{subfigure}
    \hfill
    \begin{subfigure}[b]{0.31\textwidth}
        \centering
        \includegraphics[width=\linewidth]{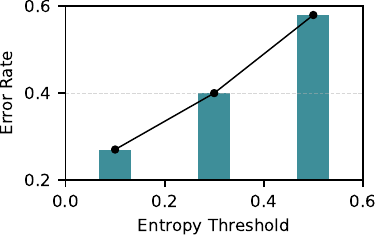}
        \vspace{-3mm}
        \caption{From FakeSV to FakeTT.}
    \end{subfigure}

    \vspace{3mm}

    \begin{subfigure}[b]{0.31\textwidth}
        \centering
        \includegraphics[width=\linewidth]{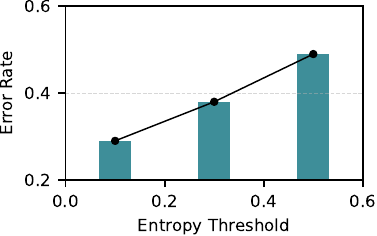}
        \vspace{-3mm}
        \caption{From FakeSV to FVC.}
    \end{subfigure}
    \hfill
    \begin{subfigure}[b]{0.31\textwidth}
        \centering
        \includegraphics[width=\linewidth]{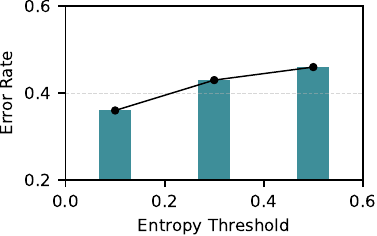}
        \vspace{-3mm}
        \caption{From FVC to FakeSV.}
    \end{subfigure}
    \hfill
    \begin{subfigure}[b]{0.31\textwidth}
        \centering
        \includegraphics[width=\linewidth]{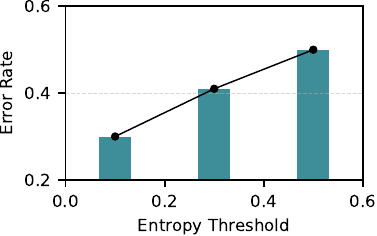}
        \vspace{-3mm}
        \caption{From FVC to FakeTT.}
    \end{subfigure}

    \vspace{-2mm}
    \caption{The relationship between the prediction entropy and the prediction accuracy across six evaluation groups.}
    \label{fig:entropy-accuracy}
    \vspace{-2mm}
\end{figure*}

\begin{figure}[t]
  \centering
  \includegraphics[width=\columnwidth]{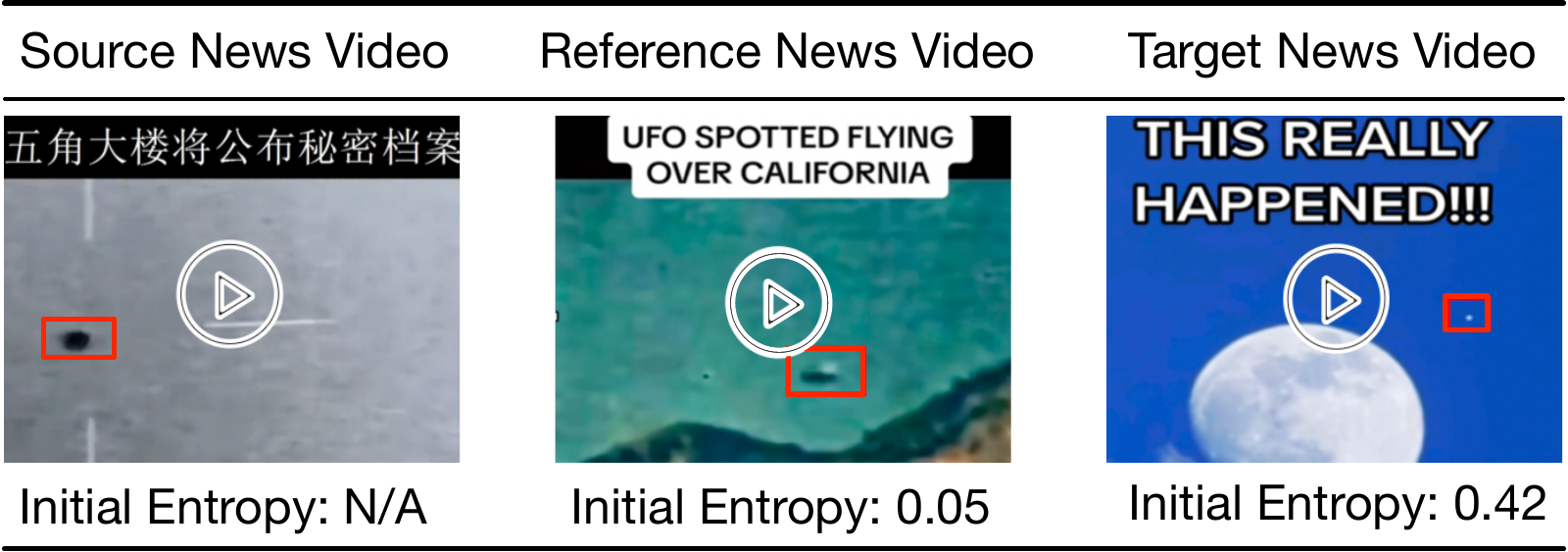}
  \vspace{-4mm}
  \caption{Case study: the stable target-domain video servers as a bridge for robust test-time adaptation.}
  \vspace{-3mm}
  \label{fig:case-study}
\end{figure}

\subsection{Empirical Validation of Low-Entropy Videos}

In prior TTA studies~\cite{wang2021tent, niu2022efficient, guo2025smoothing}, source models tend to produce more accurate predictions on target samples for which they are more confident. 
However, these TTA studies primarily address mild distribution shifts induced by synthetic corruptions (e.g., Gaussian noise, snow). 
Consequently, there is no prior empirical evidence to verify whether, in \task --- where distribution shifts are more severe and occur at the topic level --- higher confidence (i.e., low-entropy predictions) still correlates with greater prediction accuracy.
To validate this phenomenon, which serves as the cornerstone of our paradigm, we conduct an experiment to explore the relationship between prediction entropy and prediction accuracy of news videos in \task. 
Specifically, we perform cross-platform detection on six evaluation groups defined in the main paper by first training the source model on the source domain and then testing it on the target domain, without any adaptation. 
We report the prediction error rate (y-axis), i.e., $1 - \text{Acc}$, of news videos across different entropy intervals (x-axis). 
As illustrated in~\Cref{fig:entropy-accuracy}, the error rate increases with prediction entropy, indicating a negative correlation between prediction entropy and accuracy --- i.e., more confident (low-entropy) predictions tend to be more accurate in \task.

\subsection{Case Study on Retrieval-Guided Adaptation}
To further understand how our retrieval-guided adaptation functions, we conduct a case study to examine the role of low-entropy reference videos in facilitating the adaptation of unstable news videos.
We present three news videos in~\Cref{fig:case-study}: the first is from the source-domain FakeSV, while the second and third are from the target-domain FakeTT. 
The reference video (second) shares a similar fake news topic (UFO sighting) and visual content with the source video, leading to low initial prediction entropy under the source model. 
This retrieved stable reference assists the adaptation of the third, unstable (high-entropy) target video, which also involves fake news related to flying object (similar to the reference) 
but lacks obvious visual or textual similarity to the source.

\section{Conclusion}
\label{sec:conclusion}
In this study, we shed light on the challenges of test-time \task under sharp topic-level distribution shifts, including \textit{substantial representation discrepancy} and \textit{highly imbalanced label distributions}. 
We then propose \textbf{\M}, the first framework that introduces a novel retrieval-guided adaptation paradigm for effective test-time \task.
Specifically, the \textit{Entropy Selection-Based Retrieval mechanism} enhances unstable news videos with stable references.
Subsequently, the \textit{Stable Anchor-Guided Alignment module} explicitly mitigates the drastic shifts in news video representations.
Finally, the \textit{Target-Domain Aware Self-Training paradigm} assists the source model to better adapt to the ever-changing and imbalanced label distributions.
Extensive experiments conducted on three real-world video datasets 
showcase the effectiveness of our \M for test-time \task.
Future work will explore applying this retrieval-guided adaptation paradigm to other domains sharing similar situations.

\section*{Acknowledgments}
This work was supported by National Natural Science Foundation of China (Grant No. 62572097, No. 62176043, and No. U22A2097).

\bibliographystyle{ACM-Reference-Format}
\balance
\bibliography{main}

\appendix

\clearpage
\newpage
\nobalance

\section{Video Feature Extraction}
\label{app:feature}
We provide a detailed video feature extraction process, covering three modalities for each video: visual, textual, and audio. 
Specifically, for the \textbf{visual modality}, we first employ FFmpeg to uniformly sample $N_v$ frames from each video: $F = \{f_i\}_{i=1}^{N_v}$, where $N_v$ denotes the number of sampled frames ($N_v$ = 16 in this study). These frames are individually fed into a Vision Transformer (ViT)~\cite{dosovitskiyimage} (\texttt{vit-base-patch16-224}). We extract the \texttt{[CLS]} token from the final layer of each frame and perform average pooling over all frames to obtain a global visual representation: $\mathbf{v} = \frac{1}{N_v} \sum_{i=1}^{N_v} \Phi_v(f_i) \in \mathbb{R}^{D_v}$, where $D_v$ is the visual feature dimension.
For the \textbf{textual modality}, we apply Paddle-OCR~\cite{li2022ppocrv} to each sampled frame to perform optical character recognition. The recognized on-screen texts are concatenated with the video title to form the raw textual input: $T = [T_\text{title}, T_\text{screen}]$. This raw text is encoded using a multilingual BERT model~\cite{devlin2019bert} (\texttt{bert-base-multilingual-cased}) with a maximum sequence length of 128. We utilize the embedding of the \texttt{[CLS]} token as the holistic textual feature representation: $\mathbf{t} = \Phi_t(T)_{\texttt{[CLS]}} \in \mathbb{R}^{D_t}$, where $D_t$ is the textual feature dimension. 
For the \textbf{audio modality}, we extract the audio track from each video using FFmpeg and generate a transcript via Whisper~\cite{radford2023robust} (\texttt{whisper-large-v3}). 
The transcript $A$ is encoded using the same multilingual BERT model. Similarly, we take the \texttt{[CLS]} token embedding as the final audio feature representation: $\mathbf{a} = \Phi_t(A)_{\texttt{[CLS]}} \in \mathbb{R}^{D_a}$, where $D_a = D_t$ is the audio feature dimension.

\section{Detailed Experimental Setup}
\label{app:setup}
We provide a detailed version of the experimental setup, including: \textbf{(1) baselines}, \textbf{(2) datasets}, \textbf{(3) hyperparameters}, and \textbf{(4) implementation details}.

\subsection{Baselines}
To comprehensively showcase the efficacy of our proposed \M on test-time \task, we compare it with both (1) \textbf{Traditional \task methods}, including FANVM~\cite{choi2021using}, SV-FEND~\cite{qi2023fakesv}, FakRec~\cite{bu2024fakingrecipe}, ExMRD~\cite{hong2025following}, and REAL~\cite{li2025real}; and (2) \textbf{TTA methods}, including Source Model (Source), Tent~\cite{wang2021tent}, SAR~\cite{niu2023towards}, READ~\cite{yang2024testtime}, and SuMi~\cite{guo2025smoothing}.
Below, we offer a detailed description for each baseline:

\noindent (1) \textbf{Traditional \task methods}:
\begin{itemize} [leftmargin=*]
    \item \textbf{FANVM}~\cite{choi2021using} is a multimodal detection model for \task. It learns topic-invariant representations by jointly modeling stance differences between textual modalities and leveraging adversarial training to mitigate topic dependency in feature learning.
    \item \textbf{SV-FEND}~\cite{qi2023fakesv} is a multimodal detection method for fake news in short videos. It captures informative cross-modal correlations using co-attention transformers and incorporates social context signals, such as user comments and publisher profiles, to enhance detection robustness.
    \item \textbf{FakRec}~\cite{bu2024fakingrecipe} is a detection method for fake news on short video platforms that reconstructs the content generation process. It infers modality-specific intentions and captures cross-modal causal dependencies to identify inconsistencies indicative of deception.
    \item \textbf{ExMRD}~\cite{hong2025following} is an explainable detection method for news videos. 
    It formulates rumor detection as a multi-step reasoning task that explicitly models stance, consistency, and veracity to enhance interpretability and performance.
    \item \textbf{REAL}~\cite{li2025real} is a model-agnostic detection method that enhances fake news video classification by learning manipulation-aware representations. 
    It retrieves semantically relevant real and fake samples via an LLM-driven retriever and aligns the target video with category-specific prototypes to amplify subtle manipulation cues.
    We implement REAL with the basic detector SV-FEND.
\end{itemize}

\begin{table}[t]
\small
    \setlength{\tabcolsep}{5pt}
    \resizebox{\linewidth}{!}{
    \begin{tabular}{lccc}
    \toprule
    \textbf{Characteristics} & \textbf{FakeTT} & \textbf{FakeSV} & \textbf{FVC} \\
    \midrule
    Total Videos (\#) & 1,814 & 3,624 & 2,764 \\
    Fake Videos (\#) & 1,172 & 1,810 & 1,633 \\
    Real Videos (\#) & 819 & 1,814 & 1,131 \\
    Events (\#) & 286 & 738 & 305 \\
    Duration (s) & 47.69 & 39.88 & 87.83 \\
    Time Range & 19/05-24/03 & 17/10-22/02 & 16/01-18/01 \\
    Language & English & Chinese  &\makecell{English} \\
    Platform & TikTok & \makecell{ Douyin*,\\ Kuaishou} & \makecell{YouTube*,\\ Facebook,\\ Twitter} \\
    \bottomrule
    \end{tabular}
    } \vspace{1mm}
        \caption{Statistics of three datasets. * indicates the dominated platform of news videos in the dataset.}
    \label{tab:app-dataset}
    \vspace{-5mm}
\end{table}

\noindent (2) \textbf{TTA methods}:
\begin{itemize} [leftmargin=*]
    \item \textbf{Tent}~\cite{wang2021tent} is pioneering work in test-time adaptation that updates only the batch normalization parameters with entropy minimization strategy as self-supervised signals. 
    It adapts models on-the-fly using unlabeled target data, improving robustness under domain shifts without requiring source data access.
    \item \textbf{SAR}~\cite{niu2023towards} is a test-time adaptation method designed for dynamic environments. It improves adaptation stability by regularizing model updates with both temporal prediction consistency and gradient similarity across test samples, mitigating catastrophic forgetting under distribution shifts.
    \item \textbf{READ}~\cite{yang2024testtime} is a multimodal test-time adaptation approach that mitigates modality reliability bias under distribution shifts. It dynamically estimates the reliability of each modality and reweights their contributions to adaptively calibrate multimodal predictions during inference.

    \item \textbf{SuMi}~\cite{guo2025smoothing} is a multimodal test-time adaptation framework that enhances stability under complex modality-specific noises. It smooths the optimization dynamics via sample-wise regularization and adaptive feature mixing, enabling robust adaptation without source supervision.
\end{itemize}

\begin{figure*}[t]
    \centering
    \begin{subfigure}[b]{0.29\textwidth}
        \centering
        \includegraphics[width=\linewidth]{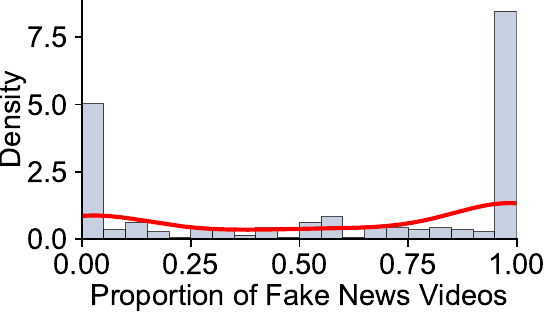}
        \vspace{-3mm}
        \caption{FakeTT Dataset.}
    \end{subfigure}
    \hfill
    \begin{subfigure}[b]{0.29\textwidth}
        \centering
        \includegraphics[width=\linewidth]{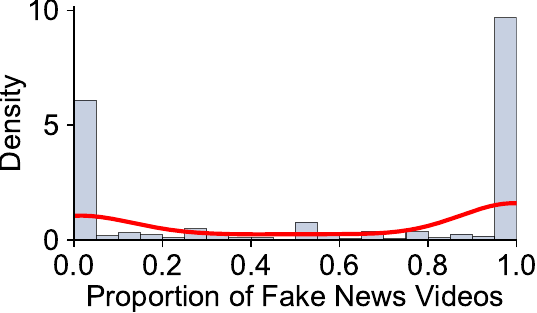}
        \vspace{-3mm}
        \caption{FakeSV Dataset.}
    \end{subfigure}
    \hfill
    \begin{subfigure}[b]{0.29\textwidth}
        \centering
        \includegraphics[width=\linewidth]{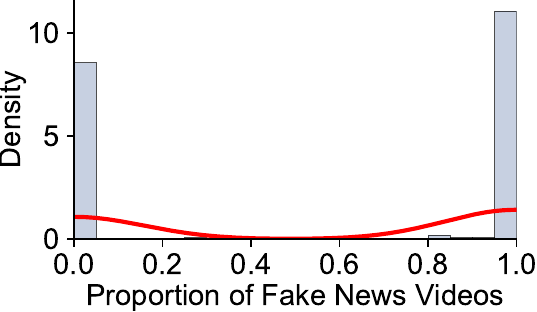}
        \vspace{-3mm}
        \caption{FVC Dataset.}
    \end{subfigure}
    \vspace{-2mm}
    \caption{The category distributions in various news events across three datasets.}
    \label{fig:cate-distribution}
    \vspace{-2mm}
\end{figure*}

\subsection{Datasets}
In this study, to simulate drastic topic-level distribution shifts in fake news videos and better reflect real-world scenarios, we adopt three datasets that differ in either event time ranges, or language and platform: FakeTT~\cite{bu2024fakingrecipe}, FakeSV~\cite{qi2023fakesv}, and FVC~\cite{papadopoulou2019corpus}, with detailed statistics provided in~\Cref{tab:app-dataset}.
During the evaluation, we conduct six groups of cross-platform detection, including \textbf{FakeTT\textrightarrow~FakeSV}, \textbf{FakeTT\textrightarrow~FVC}, \textbf{FakeSV\textrightarrow~FakeTT}, \textbf{FakeSV\textrightarrow~FVC}, \textbf{FVC\textrightarrow~FakeSV}, and \textbf{FVC\textrightarrow~FakeTT}.
Following, we present the detailed dataset descriptions:
\begin{itemize} [leftmargin=*]
\item \textbf{FakeSV}~\cite{qi2023fakesv}: FakeSV is a multimodal dataset for misinformation detection in short-form videos from Chinese platforms. It contains labeled samples collected from popular Chinese video-sharing platforms Douyin and Kuaishou, each comprising the video, title, cover image, and rich social context, including user comments and publisher metadata.

\item \textbf{FakeTT}~\cite{bu2024fakingrecipe}: FakeTT is a multimodal dataset designed for misinformation detection in English short-form videos. 
It contains labeled samples collected from TikTok, each comprising the video, title, and associated metadata such as user profiles and engagement signals.

\item \textbf{FVC}~\cite{papadopoulou2019corpus}: FVC is a multimodal dataset for fake news detection in real-world news videos. 
It consists of labeled samples collected from mainstream media YouTube and user-generated sources. 
Each sample includes video frames, audio, and automatically extracted transcripts. 

\end{itemize}

As illustrated in~\Cref{fig:cate-distribution}, we report the label distributions of fake news proportions across news events in all three datasets.
The figure shows that each dataset exhibits highly imbalanced event-level distributions, where the majority of news events contain either exclusively real or exclusively fake news videos (i.e., the proportion of fake news videos is close to either 0 or 1.).
These results can well prove our motivation to solve the unpredictable, ever-varying, and unbalanced category distributions for test-time \task.

\subsection{Hyperparameters}
\label{app:hyper}

The parameter $M$, representing the capacity of the memory bank $\mathcal{B}$ in the FIFO strategy, is set to $6 \times \text{Batch Size}$ by default.
In addition, we set the number of retrieved news videos in the multimodal similarity-based Top-$K$ retrieval in a range of $K \in [4, 12]$, and the threshold $E_0$ in the entropy selection is selected from a range of $E_0 \in [0.24, 0.6]$.
The $\alpha$ and $\beta$ follow: $\alpha + \beta = 1$, and we set both to 0.5 by default.
Finally, hyperparameter $\gamma$, utilized to balance the alignment loss $\mathcal{L}_\text{align}$, is selected from a range of $\gamma \in [10^{-1}, 10^{2}]$.

\subsection{Implementation Details}
\label{app:imp_details}
In this study, we implement the source model with a two-layer feedforward network (FFN) as each modality-specific encoder (with feature dimension as 768 for all three modalities), a two-layer transformer as the modal fusion module, and a two-layer MLP as the classifier.

For random batch sampling, the batch size $B$ is set to 128. 
For event-wise batch sampling, $B$ is set to the average number of news videos associated with a single event, in order to better simulate real-world conditions. 
Specifically, we set $B = \text{7}$ for the FakeTT dataset, $B = \text{6}$ for the FakeSV dataset, and $B = \text{9}$ for the FVC dataset.

Following prior works in TTA~\cite{niu2022efficient, guo2025smoothing}, we adopt the online adaptation setting and all the evaluation and adaptation is finished within a single epoch.
The adaptation process uses an initial learning rate of 0.0001.
For the baselines, we use their official source code and default hyperparameters, modifying only the input and output to fit the \task task and integrating them with the same source model described earlier. 
Each method is run five times, and we report the average performance as the final result.
All experiments are conducted on a system equipped with an Intel Core i9-14900KF CPU, 
an NVIDIA GeForce RTX 4090 GPU with 24 GB of VRAM, and 128 GB of system memory.

\section{Calculation of MMD Distance}
\label{app:mmd}
To quantitatively assess the severity of distribution shift caused by topic-level fake news video evolution, we adopt the Maximum Mean Discrepancy (MMD) distance~\cite{gretton2006kernel, jayasumana2024rethinking}.
Originally proposed as a statistical test for comparing two samples, MMD is widely recognized for its effectiveness in quantifying the divergence between two distributions.
Specifically, for two probability distributions $P$ and $Q$ defined over $\mathbb{R}^d$, the MMD distance with respect to a positive definite kernel $k$ is expressed as:
\begin{align}
    \operatorname{dist}^2_{\operatorname{MMD}}(P, Q) := \mathbb{E}_{\vx, \vx'}[k(\vx, \vx')] &+ \mathbb{E}_{\vy, \vy'}[k(\vy, \vy')] \nonumber  \\
    &- 2 \mathbb{E}_{\vx, \vy}[k(\vx, \vy)],
\end{align}
where $\vx$ and $\vx'$ are independently sampled from $P$, and $\vy$ and $\vy'$ are independently sampled from $Q$. It is well-established that MMD serves as a valid metric for characteristic kernels $k$~\cite{fukumizu2008characteristic}.
Given two sets of vectors, $X = \{\vx_1, \vx_2, \dots, \vx_m\}$ and $Y = \{\vy_1, \vy_2, \dots, \vy_n\}$, drawn from $P$ and $Q$ respectively, an unbiased estimator for $d^2_{\operatorname{MMD}}(P, Q)$ is defined as follows:
\begin{align}
    \hat{\operatorname{dist}}_{\operatorname{MMD}}^2(X, Y) =& \frac{1}{m(m-1)}\sum_{i=1}^m\sum_{j \ne i}^m k(\vx_i,\vx_j) \nonumber \\
    &+ \frac{1}{n(n-1)}\sum_{i=1}^n\sum_{j \ne i}^n k(\vy_i,\vy_j) \nonumber \\
    &- \frac{2}{mn}\sum_{i=1}^m\sum_{j=1}^n k(\vx_i, \vy_j).
    \label{eqn:mmd_from_a_sample}
\end{align}
In this study, we use the Gaussian RBF kernel $k(\vx, \vy) = \exp(-\|\vx - \vy\|^2/2\sigma^2)$, which is a characteristic kernel,  with the bandwidth parameter set to $\sigma=1.0$. 
As MMD operates on distributions over $D$-dimensional feature vectors, we convert each $S \times D$ input into a single D-dimensional representation.
Specifically, we average the frame-level features for vision inputs and use the [CLS] token representation for both text and audio inputs.

\end{document}